\documentclass{article} 
\usepackage{iclr2027_conference,times}
\def\iclrruler#1{}   

\usepackage{amsmath}
\usepackage{amssymb}
\usepackage{amsthm}
\usepackage{etoolbox}
\makeatletter
\newcommand{\eqsize}{\edef\@eqsavedbls{\the\baselineskip}\small\baselineskip=\@eqsavedbls\relax}
\newcommand{\nonfloatfigurecaption}[1]{\def\@captype{figure}\caption{#1}}
\AtBeginEnvironment{equation}{\eqsize}
\AtBeginEnvironment{equation*}{\eqsize}
\def\maketag@@@#1{\hbox{\m@th\normalsize\normalfont#1}}
\makeatother

\usepackage{graphicx}
\usepackage{booktabs}
\usepackage{float}   
\usepackage{wrapfig}
\usepackage{listings}
\usepackage[utf8]{inputenc}
\usepackage[T1]{fontenc}   
\usepackage{hyperref}
\usepackage{url}

\title{Learning to Think and Explore in 3D World}

\author{%
Yijiang Li$^{1,\dagger}$ \quad Bingyang Wang$^{2,\dagger}$ \quad Nuno Vasconcelos$^{1}$ \\[3pt]
\normalfont $^{1}$University of California San Diego
\normalfont $^{2}$Georgia Institute of Technology \\[2pt]
\normalfont\texttt{yijiangli@ucsd.edu}%
}
\newcommand{\blfootnote}[1]{%
  \begingroup\renewcommand\thefootnote{$\dagger$}\footnotetext{#1}\endgroup}

\newcommand{\best}[1]{\textbf{#1}}

\usepackage{multirow}
\usepackage{colortbl}
\definecolor{indomain}{RGB}{214,240,214}
\newcommand{\indom}[1]{\cellcolor{indomain}#1}
\definecolor{oursrow}{gray}{0.92}
\definecolor{blockblue}{RGB}{0,72,153}
\newcommand{\blockhdr}[1]{\textcolor{blockblue}{#1}}
\definecolor{sysbg}{HTML}{EEF1FD}
\definecolor{usrbg}{HTML}{E9F5EC}
\usepackage{framed}                       
\definecolor{shadecolor}{HTML}{F0F0F0}    
\newcommand{\tg}[1]{\texttt{\textless #1\textgreater}}
\lstdefinestyle{promptbox}{
  basicstyle=\scriptsize\ttfamily,
  breaklines=true,
  breakindent=0pt,
  columns=fullflexible,
  keepspaces=true,
  upquote=true,
  frame=none,
  xleftmargin=4pt,
  xrightmargin=4pt,
  aboveskip=3pt,
  belowskip=0pt,
}
\begin{document}

\maketitle
\blfootnote{Equal contribution.}

\begin{abstract}

We study how vision-language models (VLMs) can learn to think and actively explore in 3D space. Instead of providing multi-view images, videos, or 3D representations to the model, we preserve the native visual encoding of VLMs, i.e., a single image, one at a time, just like humans, and treat the 3D representation only as an environment with which the model can interact through actions (e.g., move left) and observations (rendered images). Each rollout begins from an initial first-person view and a visual question. The model can then iteratively reason, navigate, and reorient its viewpoint before committing to a final answer. Following an initial trajectory-supervised finetuning, we apply reinforcement learning (RL) with rewards driven primarily by final-answer correctness. Crucially, RL provides no direct supervision over which viewpoints the model should visit or which exploration actions it should take. Instead, the model must learn an effective exploration policy solely from the extent to which its interactions improve performance. Across six 3D benchmarks, our model ACT3D achieves an average accuracy of 53.2\% that improves its base Qwen model by 22.0\% and surpasses proprietary Gemini 3.6 Flash. ACT3D also outperforms existing open-source 3D methods while using 50-70\% fewer input frames, and generalizes to unseen domains with an average improvement of 5.9\% over its supervised initialization. Further analysis reveals adaptive, question-conditioned exploration. Performance gains over the base model increase from 0.6\% on questions answerable from the initial view to 16.3\% on those requiring extensive exploration. ACT3D searches for occluded targets, avoids unnecessary exploration, and acquires complementary viewpoints for spatial comparisons. These findings demonstrate that outcome-based RL enables VLMs not only to reason over visual evidence, but also to learn when and how to acquire it. Our project page is \url{https://williamium3000.github.io/act3d/}.

\end{abstract}

\begin{figure}[h]
    \centering
    \includegraphics[width=1\linewidth]{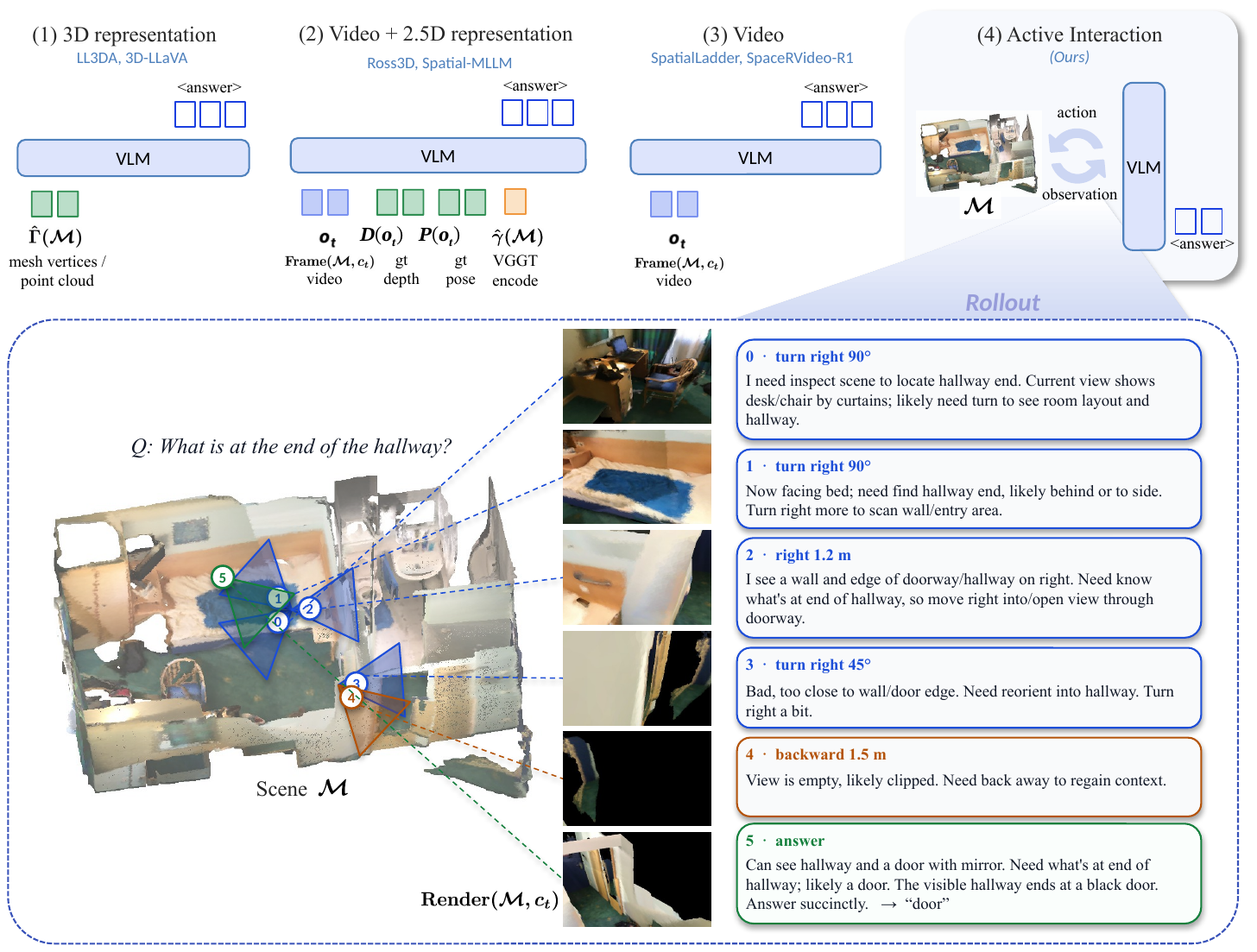}
    \caption{\textbf{Learning to reason and explore in 3D.} Unlike prior methods with passive perception which assumes receive complete 3D representations as input, Act3D treats 3D representations as interactive environments, learning to alternate between reasoning and actions to actively explore the 3D world before committing an answer.}
    \label{fig:teaser}
\end{figure}


\vspace{-3mm}

\section{Introduction}

Despite substantial progress in 3D vision-language models, much of the existing work studies spatial reasoning under observations that are specified \emph{a priori}, providing models with complete RGB-D scans~\citep{azuma2022scanqa,zhu2025llava3d}, multi-view images~\citep{yang2025thinking}, videos~\citep{zheng2025video3dllm,yang2026cambrians}, or explicit 3D scene representations such as point clouds and meshes \citep{hong2023threedllm,huang2024leo}.
Yet under partial observability, a fundamental challenge remains: the model must not only reason over the evidence currently available, but also determine how to act in order to acquire the evidence that is missing. This motivates the central question of our work: \emph{can a vision-language model learn to actively explore and acquire task-relevant visual evidence as an integral part of the 3D reasoning process?}

We address this question by first formulating reasoning in 3D as an interactive visual decision process under partial observability. Rather than providing the VLM with a complete 3D representation, the 3D scene is kept external to the VLM and serves as an interactive environment that the model accesses through sequential action-observation interactions. Starting from a fixed initial egocentric view and a visual question, the model alternates between reasoning and interaction, iteratively acquiring observations through navigation and viewpoint control before committing to a final answer. This entire episode of interaction constitutes a policy rollout, in which the model jointly determines \emph{what information to acquire, how to interact with the environment, and how to reason over the accumulated evidence}.
To learn a policy that integrates visual information acquisition, spatial interaction, and reasoning, we adopt a two-stage training procedure: trajectory-supervised fine-tuning first establishes the interaction protocol, after which reinforcement learning optimizes the full interactive policy over the rollout described above using rewards driven primarily by final-answer correctness. During reinforcement learning, no direct supervision is provided for intermediate viewpoints, actions, or exploration trajectories. Effective exploration must instead be learned through credit assignment from task performance. This formulation enables us to examine whether task-level supervision alone can induce task-directed visual information acquisition during 3D reasoning.

Building on this formulation, we present Act3D, a vision-language model that jointly learns to acquire visual evidence and reason over observations. Experiments across six 3D benchmarks show that Act3D achieves an average accuracy of 53.2\%, outperforming all evaluated prior methods and baselines. It improves upon its base Qwen model by 22.0\% and surpasses Gemini 3.6 Flash by 1.0\%. Act3D also outperforms existing open-source 3D methods while using 50--70\% fewer input frames and lower-resolution observations. These gains generalize to out-of-distribution beyond the training domains, with an average improvement of 5.9\% over the supervised initialization across four out-of-domain benchmarks. Beyond accuracy, we observe the emergence of adaptive exploration behaviors where models learn to explore more with increasing interaction steps on questions that require more exploration, leading to pronounced improvements when target objects are initially occluded or far away from the initial camera position. We further qualitatively illustrate that Act3D learns to search for hidden objects, limit unnecessary exploration when relevant evidence is already visible, and gather complementary viewpoints to resolve spatial relationships.
Together, these findings demonstrate that RL enables VLMs not only to reason over visual observations, but also to adaptively acquire the evidence needed for 3D reasoning.

\vspace{-2mm}
\section{Related Work}
\label{sec:related}

\vspace{-2mm}


Prior work on 3D vision-language models operates over observations assumed to be complete and directly presented to the model, including 3D representations such as point clouds and object-centric features~\citep{hong2023threedllm,huang2024chatscene,huang2024leo,jin2023context,zhu2023vista,chen2024ll3da,fu2025scenellm,zhu2024pq3d,chen2024grounded3dllm,kang2025robin3d}, fixed multi-view images or videos~\citep{zhu2025llava3d,zheng2025video3dllm,qi2025gpt4scene}, and representations enhanced through reconstruction supervision~\citep{wang2025ross3d}.
A parallel line of work improves spatial reasoning in general-purpose VLMs through large-scale spatial supervision~\citep{chen2024spatialvlm,yang2026cambrians} and reinforcement learning with verifiable rewards~\citep{ouyang2025spacer,wu2025vilasr,li2025spatialladder}.
Despite advances in spatial reasoning, these approaches largely assume that the visual evidence required to answer a question is provided to the model, rather than actively acquired during inference.
In contrast, Act3D keeps the 3D scene representation external to the VLM as an interactive environment: the model receives only egocentric 2D observations and learns to navigate and acquire task-relevant visual evidence through interaction, jointly with reasoning.

\textbf{Multimodal agentic reasoning.}

Reinforcement learning (RL) (e.g.,  GRPO \citep{shao2024deepseekmath, wang2026learn}) has emerged as an effective paradigm for improving reasoning capabilities~\citep{guo2025deepseekr1, yang2026co, yu2025dapo, zhao2026self, li2026policy}. It has further been applied to agentic systems with multi-turn interaction and tool use~\citep{jiang2026verltool, li2026contextualized, wang2025ragen, jin2025searchr1, luo2025agentlightning}, as well as to multimodal reasoning~\citep{feng2025videor1, feng2025videoorion, li2026self, li2024core, huang2025visionr1, liu2025visualrft, shen2025vlmr1}.
These methods primarily manipulate or revisit visual evidence already contained in the provided input. In contrast, our model learns to explore the 3D environment, actively acquiring new visual evidence while jointly optimizing reasoning and interaction in a single rollout.

\textbf{Exploration in embodied and 3D environments.} EmbodiedQA introduced navigation-based question answering \citep{das2018embodied}, later extended to open-vocabulary evaluation by OpenEQA \citep{majumdar2024openeqa}.
Recent benchmarks further show that active observation can outperform passive viewing and expose failures in spatial exploration and long-horizon planning \citep{esibench2026,zhang2026vabench,zhang2026theoryofspace,wei2026seerememberexplore,wang2026planningviews}. We abstract embodiment to direct camera control over a scanned 3D scene and jointly learn exploration and reasoning for question answering.
More closely related are methods that actively select viewpoints. Chain-of-View performs training-free coarse-to-fine view selection through a discrete camera-action interface \citep{zhao2026cov}. Think3D equips models with 3D reconstruction tools and optimizes their use with RL \citep{zhang2026think3d}. ViewSuite distills self-exploration into a view graph \citep{wang2026planningviews}.


\vspace{-2mm}
\section{Method}
\label{sec:method}

We study how a VLM can learn to reason in 3D under partial observability.
Learning such a policy poses two complementary challenges.
The model must first acquire a basic interaction competence in 3D environments before it can learn from the interaction with environments despite the absence of ground-truth exploration trajectories. 
We therefore adopt a two-stage training procedure. We first perform \emph{trajectory-supervised fine-tuning} using curated interaction demonstrations to initialize a coherent reason--act--observe policy. We then apply \emph{reinforcement learning with verifiable reward} to optimize the policy with rewards driven only by final performance, without supervision of any intermediate viewpoints or actions. This allows the model to learn task-directed exploration end-to-end, supervised only by how much the evidence it acquires improves its answers.

\vspace{-2mm}
\subsection{Reasoning in 3D under Partial Observability}
\label{sec:pomdp}

\textbf{Problem formulation.}
We formalize reasoning in 3D as a partially observable Markov decision process (POMDP)
\citep{kaelbling1998planning}.
A task instance $w=(\mathcal{M},q,y^\star)\sim\mathcal{D}$ consists of a reconstructed 3D scene $\mathcal{M}$, a question $q$, and the ground-truth answer $y^\star=g(\mathcal{M},q)\in\mathcal{Y}$. At step $t$, the environment state is $s_t=(\mathcal{M},c_t)$, where $c_t=(e_t,f_t)$ specifies the camera position $e_t\in\mathbb{R}^3$ and viewing direction $f_t\in\mathbb{S}^2$, and $c_0$ is a fixed initial viewpoint of the scene. We write $\hat z$ for the vertical axis of the world frame; the camera up vector is held parallel to $\hat z$, so the camera carries no roll and $c_t$ has five degrees of freedom. The scene $\mathcal{M}$ remains fixed and hidden throughout the episode, while the camera state evolves under the model's actions. The policy observes the rendered egocentric view $o_t=\operatorname{Render}(\mathcal{M},c_t)$ and, after each action, the resulting camera configuration $c_t$.



\textbf{Action space.}
We consider two parameterizations of how the policy specifies the next viewpoint.

Under the \emph{relative} parameterization, a navigation action
$a_t=(\xi_t,n_t)\in\mathcal{A}_{\mathrm{nav}}=\mathcal{U}\times\{1,\ldots,8\}$ pairs a motion
primitive $\xi_t$ with an integer multiplier $n_t$, giving $|\mathcal{A}_{\mathrm{nav}}|=80$
navigation actions. The set $\mathcal{U}$ contains six translations
(\texttt{forward}/\texttt{backward}, \texttt{left}/\texttt{right}, \texttt{up}/\texttt{down}) and
four rotations (\texttt{turn\_left}/\texttt{turn\_right} and
\texttt{turn\_up}/\texttt{turn\_down}). Each primitive has a unit step $\delta_\xi$, equal to
$1$\,m for \texttt{forward}/\texttt{backward}, $0.6$\,m for the remaining translations, and
$30^\circ$ for the rotations. The policy may additionally supply a step size $d$, in metres for
translations and degrees for rotations, which the environment quantizes to the multiplier
$n_t=\operatorname{clip}(\operatorname{round}(d/\delta_{\xi_t}),1,8)$, so that a single action covers
at most eight unit steps; omitting $d$ gives $n_t=1$. Let
$r_t=(f_t\times\hat z)/\|f_t\times\hat z\|$ denote the camera's lateral direction, which is
horizontal for any non-vertical $f_t$. Camera dynamics are deterministic:
\begin{equation}
(e_{t+1},f_{t+1})
=
\begin{cases}
\left(\Pi_{\mathcal B}(e_t+n_t\delta_{\xi_t}u_t),\,f_t\right),
& a_t \text{ is a translation},\\[1mm]
\left(e_t,\,\operatorname{Rot}_{k_t}(n_t\delta_{\xi_t})f_t\right),
& a_t \text{ is a rotation},
\end{cases}
\end{equation}
where $\operatorname{Rot}_{k}(\alpha)$ rotates a vector by an angle $\alpha$ about the unit axis
$k$. The primitive $\xi_t$ selects the translation direction $u_t$ or the rotation axis $k_t$:
translations use $u_t=+f_t,-f_t,+r_t,-r_t,+\hat z,-\hat z$ for \texttt{forward}, \texttt{backward},
\texttt{right}, \texttt{left}, \texttt{up} and \texttt{down}, and rotations use $k_t=+\hat z,-\hat z$
for \texttt{turn\_left}/\texttt{turn\_right} and $k_t=+r_t,-r_t$ for
\texttt{turn\_up}/\texttt{turn\_down}. Because $r_t$ is built from $\hat z$, lateral motion stays
horizontal and the horizon stays level at every pose. Finally, $\mathcal B$ is the axis-aligned
bounding box of the scene enlarged by $1$\,m along each axis, and $\Pi_{\mathcal B}$ is the Euclidean
projection onto it, which acts as a componentwise clamp; it keeps the camera near the scene but
performs no collision checking, so the camera may pass through scene geometry.

Under the \emph{absolute} parameterization, an action instead names a target pose directly, as a
camera position together with a look-at point, $a_t=(p_t,\ell_t)\in\mathbb{R}^3\times\mathbb{R}^3$,
and the camera is placed at that pose:
\begin{equation}
e_{t+1}=p_t,
\qquad
f_{t+1}=\frac{\ell_t-p_t}{\|\ell_t-p_t\|}.
\end{equation}

We adopt the relative parameterization throughout training and evaluation. Absolute actions require
the policy to predict coordinates in a global reference frame that is not directly observed and whose
scale varies across scenes. In contrast, relative actions are scale-invariant, camera-centric, and
map naturally to interpretable motions such as ``move left'' or ``turn right''. We compare the two
parameterizations empirically in Appendix~\ref{sec:abl:actionspace}.

We augment navigation with the terminal answer actions
$\mathcal{A}_{\mathrm{stop}}=\{\textsc{stop}(y):y\in\mathcal{Y}\}$,
\begin{equation}
\bar{\mathcal A}
=
\mathcal A_{\mathrm{nav}}
\cup
\mathcal A_{\mathrm{stop}}.
\end{equation}
Executing $\textsc{stop}(y)$ terminates the episode and returns $y$ as the prediction. Given the task
instance, both the transition and observation processes are deterministic; stochasticity arises only
from the policy itself.

\textbf{History and implicit belief.}
At step $t$, the policy conditions on the complete interaction history
\begin{equation}
h_t=
(q,o_0,z_0,a_0,c_1,o_1,z_1,\ldots,a_{t-1},c_t,o_t,z_t),
\end{equation}
where $z_t$ denotes the model's intermediate reasoning before taking the next action. The policy is therefore $a_t\sim\pi_\theta(\cdot\mid h_t),
\
a_t\in\bar{\mathcal A}.$
Because the scene is hidden but static, the posterior
$
b_t(\mathcal M)
=
\Pr(\mathcal M\mid h_t)
$,
together with the observed camera configuration $c_t$ and the elapsed step $t$, forms a sufficient statistic for optimal decision making. In practice, we do not explicitly maintain $b_t$; instead, the VLM conditions on the full interaction context, which implicitly encodes its belief over the accumulated observations. The reasoning $z_t$ is generated by the policy and belongs to the history, not to the environment state.

\textbf{Objective.}
An episode terminates when the policy emits an answer or reaches the interaction budget, with
$
T
=
\min\{t:a_t\in\mathcal A_{\mathrm{stop}}\}
\wedge
T_{\max}$;
if no answer is emitted within the budget, the prediction $y$ is empty and counted as incorrect.
The resulting rollout is $\tau=(h_T,y)$,
and we optimize the policy for expected finite-horizon, undiscounted return:
\begin{equation}
\max_\theta J(\theta)
=
\mathbb E_{w\sim\mathcal D}
\mathbb E_{\tau\sim\pi_\theta\otimes\mathcal E}
\left[
R(\tau)
\right],
\end{equation}
where $\mathcal E$ denotes the deterministic interactive environment defined above and $R$ is the reward introduced in Sec.~\ref{sec:method-rl}, which depends on the task instance through $y^\star$ and the annotated answer object. We define the answer accuracy of a policy $\pi$ as
\begin{equation}
\operatorname{Acc}(\pi)
=
\Pr_{w\sim\mathcal D,\,
\tau\sim\pi\otimes\mathcal E}
(y=y^\star).
\end{equation}
This objective jointly evaluates the policy's ability to acquire informative observations and reason over them to produce the correct answer.

We illustrate the necessity of interaction with environment for reasoning under partial observability in Appendix~\ref{app:interaction}.


\begin{figure}[t]
\small
\hrule height0.9pt
\vspace{4pt}

\textbf{System.}\enspace 3D spatial reasoning assistant exploring an indoor scene, given ONE
starting view. Call \texttt{render\_view\_tool} with a single action --- \texttt{forward},
\texttt{backward}, \texttt{left}, \texttt{right}, \texttt{up}, \texttt{down} or
\texttt{turn\_\{left,right,up,down\}} --- and an optional distance. Answer only when confident.
Respond in exactly one of two formats:
\begin{lstlisting}[style=promptbox]
<think>...</think> <tool_call>{"action": ..., "distance": ...}</tool_call>
<think>...</think> <answer>...</answer>
\end{lstlisting}

\vspace{4pt}
\textbf{User, turn 0.}\enspace \tg{image}, the room's approximate extent,
\texttt{Question: \{question\}}, and an answer limit of 1--5 words.

\vspace{4pt}
\hrule height0.9pt
\caption{\textbf{System and User Prompt template.} See details in Appendix~\ref{app:policy-prompt}.}
\label{fig:prompt-text}
\end{figure}

\subsection{Trajectory-Supervised Policy Initialization}
\label{sec:method-sft}

\begin{figure}
    \centering
    \includegraphics[width=1\linewidth]{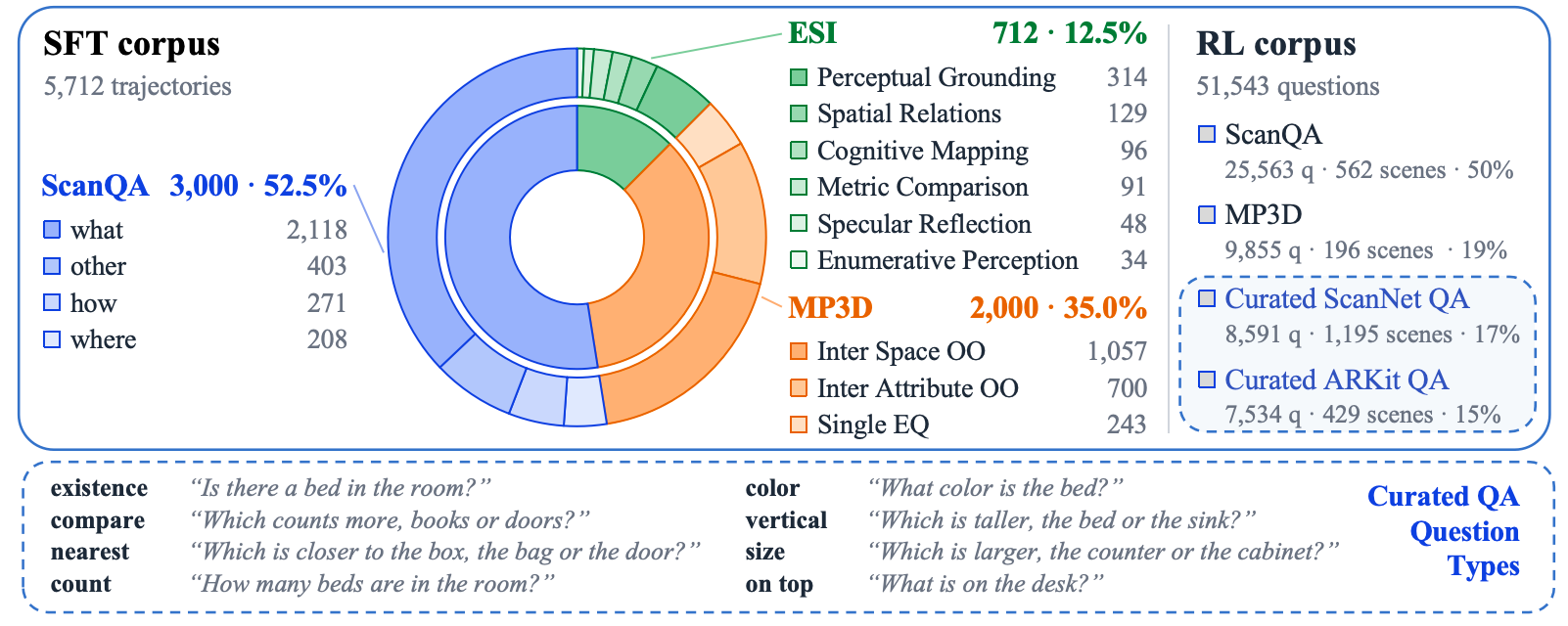}
    \caption{Training data composition. Distribution of SFT trajectories and RL questions across datasets, with representative examples of curated spatial reasoning questions.}
    \label{fig:data}
\end{figure}

We initialize the policy with trajectory-supervised fine-tuning to establish basic multi-step interaction competence in 3D environments before policy optimization with reinforcement learning.

\textbf{Trajectory curation.}
We use GPT-5.4 to generate reason--act--observe trajectories in the environment from
Sec.~\ref{sec:pomdp}. We discard invalid starting views and retain trajectories with correct
answers, valid interactions, well-formed reasoning and responses, and sufficient visible scene
geometry; MP3D examples must also observe the evidence required by the question. After answer
canonicalization, the corpus contains $5{,}712$ trajectories, $20{,}603$ rendered observations,
and $830$ scenes. These demonstrations initialize the interaction protocol rather than prescribe
a unique exploration strategy.

\textbf{Supervised objective.}
Let $\tau^\star\in\mathcal D_{\mathrm{traj}}$ denote a curated demonstration, serialized as a token sequence, and let $\mathcal I(\tau^\star)$ denote the positions of its model-generated tokens, i.e., reasoning, actions, and the final answer. Environment observations and prompt tokens are masked from the loss. We minimize
\begin{equation}
\mathcal L_{\mathrm{SFT}}(\theta)
=
-\mathbb E_{\tau^\star\sim\mathcal D_{\mathrm{traj}}}
\sum_{\ell\in\mathcal I(\tau^\star)}
\log
\pi_\theta
\left(
\tau^\star_\ell\mid \tau^\star_{<\ell}
\right).
\end{equation}
The resulting policy initializes the reinforcement learning stage described next.
We emphasize that these trajectories provide \emph{policy initialization} rather than canonical exploration targets. Trajectory supervision therefore establishes the interaction protocol and a useful behavioral prior, while the acquisition strategy is subsequently optimized with RL.
\vspace{-2mm}
\subsection{Outcome-Driven Reinforcement Learning}
\label{sec:method-rl}

Starting from the trajectory-supervised initialization, we optimize the full interactive policy using outcome-driven reinforcement learning. Each rollout follows the decision process in Sec.~\ref{sec:pomdp}.

\textbf{Training data.}
RL requires only a question--answer pair and a valid initial viewpoint, allowing substantially broader training data than trajectory supervision. We collect $51{,}543$ questions from ScanQA, MP3D, and generated QAs over ScanNet and ARKitScenes.

For each initial condition $x=(q,o_0)$, we sample a group of $G$ rollouts $\{\tau_i\}_{i=1}^{G}$ from $\pi_{\theta_{\mathrm{old}}}\otimes\mathcal{E}$, all starting from $x$.
Because actions determine subsequent observations, rollouts from the same initial condition may acquire substantially different evidence, enabling direct comparison of alternative information-acquisition strategies under the same task objective.

\textbf{Verifiable Reward.}
For a rollout $\tau$ with final answer $y$, the reward is a weighted sum of answer-correctness, interaction, and output-format terms:
\begin{equation}
R(\tau)
=
\underbrace{\lambda_{\mathrm{judge}}R_{\mathrm{judge}}+\lambda_{\mathrm{EM}}R_{\mathrm{EM}}}_{\text{answer correctness}}
+
\underbrace{\lambda_{\mathrm{int}}R_{\mathrm{judge}}\,c(\tau)}_{\text{interaction reward}}
-
\underbrace{\lambda_{\mathrm{fmt}}R_{\mathrm{fmt}}}_{\text{output format}},
\end{equation}
where $\lambda_{(\cdot)}\geq 0$ are specified, together with $N_{\max}$ in Sec.~\ref{sec:setup}. All terms depend on $\tau$; we omit this for brevity. $R_{\mathrm{judge}}\in\{0,1\}$ is the verdict of an LLM judge that, given the question, determines whether $y$ is semantically equivalent to the ground-truth answer $y^\star$. $R_{\mathrm{EM}}\in\{0,1\}$ indicates whether $y$ exactly matches any of the annotated GT answers; it complements the judge by favoring the concise canonical answers on which the benchmark metric is computed. The interaction reward is credited only in rollouts that end in a correct answer, and it treats verified grounding as a shortcut to the full credit rather than as a separate bonus:
\begin{equation}
c(\tau)=
\begin{cases}
N_{\max}, & R_{\mathrm{gnd}}=1,\\
\min\{n_{\mathrm{int}},N_{\max}\}, & \text{otherwise,}
\end{cases}
\end{equation}
where $n_{\mathrm{int}}$ is the number of interactions in $\tau$ and $R_{\mathrm{gnd}}\in\{0,1\}$ uses the annotated 3D location of the object of interest, taking the value $1$ when that object is visible in some rendered view of $\tau$. A rollout that is judged correct \emph{and} has actually brought the asked-about object into view therefore receives the whole $\lambda_{\mathrm{int}}N_{\max}$ at once, after as little as a single call, while a correct rollout that never shows the object still accrues credit per call up to the same cap. The intent is to pay for evidence rather than for looking: once the needed view has been obtained, further calls earn nothing. Finally, $R_{\mathrm{fmt}}\in\{0,1\}$ indicates a malformed output, namely a missing or empty answer. None of these terms prescribes a particular intermediate viewpoint, action, or trajectory; the grounding term only verifies whether the answer object was observed.

\paragraph{Policy optimization.}
We optimize the policy with Group Relative Policy Optimization (GRPO)~\citep{shao2024deepseekmath}. For each rollout group, we standardize the outcome reward to obtain
\begin{equation}
\hat A_i
=
\frac{
R(\tau_i)-\operatorname{mean}_{j}R(\tau_j)
}{
\operatorname{std}_{j}R(\tau_j)
}.
\end{equation}
The same trajectory-level advantage $\hat A_i$ is assigned to all model-generated tokens in $\tau_i$, while environment observations are excluded from optimization. GRPO then optimizes
\begin{equation}
\begin{aligned}
\mathcal J_{\mathrm{GRPO}}(\theta)
={}&
\mathbb E_{w\sim\mathcal D,\,\{\tau_i\}\sim\pi_{\theta_{\mathrm{old}}}\otimes\mathcal E}
\Bigg[
\frac{1}{G}
\sum_{i=1}^{G}
\frac{1}{|\mathcal I(\tau_i)|}
\sum_{\ell\in\mathcal I(\tau_i)}
\min
\Bigg(
\frac{\pi_\theta(\tau_{i,\ell}\mid \tau_{i,<\ell})}{\pi_{\theta_{\mathrm{old}}}(\tau_{i,\ell}\mid \tau_{i,<\ell})}\hat A_i,\\
&\qquad
\operatorname{clip}
\bigg(\frac{\pi_\theta(\tau_{i,\ell}\mid \tau_{i,<\ell})}{\pi_{\theta_{\mathrm{old}}}(\tau_{i,\ell}\mid \tau_{i,<\ell})},1-\epsilon,1+\epsilon\bigg)\hat A_i
\Bigg)
\Bigg]
-
\beta
D_{\mathrm{KL}}
\left(
\pi_\theta
\,\middle\|\,
\pi_{\mathrm{ref}}
\right),
\end{aligned}
\end{equation}
where $\mathcal I(\tau_i)$ denotes the model-generated token positions of $\tau_i$ and $\tau_{i,<\ell}$ its preceding tokens, including the rendered observations, $\epsilon$ is the clipping range, and $\beta$ weights a KL penalty toward a reference policy $\pi_{\mathrm{ref}}$.
Because reasoning and interaction are generated by the same autoregressive policy, the episode-level reward assigns credit across the entire rollout. The policy can therefore learn not only how to reason over acquired evidence, but also which evidence to acquire in the first place.

\vspace{-2mm}
\section{Experiment}
\vspace{-2mm}
\subsection{Setup}
\label{sec:setup}

\textbf{Models and training.}

All experiments use Qwen3.5-9B \citep{qwen3.5} as the base model. We consider two training configurations: ScanQA-only and mixed-data. 
In the ScanQA-only setting, SFT uses 2,550 teacher trajectories, followed by GRPO on 25,563 ScanQA questions. In the mixed-data setting, SFT uses 5,712 trajectories from ScanQA~\citep{azuma2022scanqa}, MP3D~\citep{chang2017matterport3d}, and ESI~\citep{esibench2026}, while GRPO is performed on 51,543 questions from ScanQA, MP3D, and curated QA data from ScanNet and ARKitScenes~\citep{baruch2021arkitscenes}. Both settings follow the same two-stage training procedure, and we evaluate base model, SFT initialization, and RL-trained policy to isolate the contribution of each training stage.

For SFT, we train for two epochs with a batch size of 16 and a learning rate of $10^{-5}$. GRPO uses 64 prompts per step, a mini-batch size of 16, and $G=8$ rollouts per prompt, with a learning rate of $2\times10^{-6}$, sampling temperature of $0.7$, clipping parameter $\epsilon=0.2$, and KL coefficient $\beta=0.005$, without entropy regularization. We set $(\lambda_{\mathrm{judge}}, \lambda_{\mathrm{EM}}, \lambda_{\mathrm{fmt}}, \lambda_{\mathrm{int}})=(0.3,0.2,0.2,0.05)$ and $N_{\max}=12$. A correct rollout that observes the annotated answer object receives the maximum interaction reward of $\lambda_{\mathrm{int}}N_{\max}=0.60$, whereas one that does not observe the object accumulates the reward per interaction, up to the same limit. Training is conducted on NVIDIA H200 GPUs with asynchronous rollouts using vLLM, a context length of $24{,}576$ tokens, and a maximum of twelve interaction steps per episode. Prompt and environment-observation tokens are excluded from the training loss. Full prompts are provided in Appendices~\ref{app:policy-prompt} and~\ref{app:sft-teacher}.


\textbf{Evaluation.}
We evaluate all models under a unified interactive protocol. Each model is initialized with a single $640\times480$ render of the scene and is allowed up to twelve interactions to acquire additional observations; all active baselines are evaluated using the same interaction harness and budget. We use Qwen3-30B-A3B-Instruct \citep{qwen3technicalreport} as the answer judge throughout evaluation. We evaluate on six benchmarks spanning diverse indoor environments and spatial reasoning settings: SPBench-SI and SPBench-MV \citep{li2025spatialladder}, ARKitScenes \citep{baruch2021arkitscenes}, ScanQA \citep{azuma2022scanqa}, SQA3D \citep{ma2023sqa3d}, and MP3D \citep{chang2017matterport3d}. Together, these benchmarks cover settings ranging from single-room visual question answering to multi-scene spatial reasoning.

We compare Act3D with a broad range of existing 3D reasoning methods and general-purpose VLMs. For passive perception, we include Video-R1 \citep{feng2025videor1}, ViLaSR \citep{wu2025vilasr}, SpatialLadder \citep{li2025spatialladder}, SpaceR \citep{ouyang2025spacer}, and Ross3D \citep{wang2025ross3d}, alongside InternVL3.5-38B \citep{internvl3.5} evaluated with fixed visual observations. For active interaction, we evaluate open-source VLMs, including Qwen2.5-VL-72B \citep{qwen2.5vl}, Qwen3.5-9B/27B \citep{qwen3.5}, and InternVL3.5-38B, as well as the proprietary Gemini 3.6 Flash \citep{gemini3.6flash}. These models are evaluated under the same observation--action interface and interaction budget, whereas passive methods retain their respective input and inference protocols. We additionally report the base model and SFT checkpoints to isolate the contributions of different training stages.

\vspace{-2mm}
\subsection{Main Results}
\label{sec:result}

\begin{table}[h]
\centering
\small
\setlength{\tabcolsep}{4pt}
\begin{tabular}{lccccccc}
\toprule
model & ScanQA & SQA3D & SPB MV & SPB SI & MP3D & ARKit & Avg. \\
\midrule
\multicolumn{8}{c}{\blockhdr{\emph{Passive perception}}} \\
\midrule
Video-R1-7B & 30.8 & 41.1 & 42.0 & 39.6 & 42.6 & 63.4 & \textit{43.2} \\
ViLaSR-7B & 33.3 & 47.6 & \underline{52.6} & 43.4 & 34.6 & 66.0 & \textit{46.2} \\
SpatialLadder-3B & 29.3 & 40.4 & \indom{\textbf{61.2}} & \indom{\textbf{45.2}} & 45.2 & 58.7 & \textit{46.7} \\
InternVL3.5-38B & 36.2 & 47.0 & 36.9 & 38.7 & 51.1 & \underline{75.4} & \textit{47.6} \\
SpaceR-7B & 31.6 & 44.7 & \underline{52.6} & 41.0 & 52.0 & 70.1 & \textit{48.7} \\
Ross3D & \indom{\textbf{50.7}} & \indom{\textbf{62.6}} & 32.0 & 31.9 & 55.2 & 71.5 & \textit{50.7} \\
\midrule
\multicolumn{8}{c}{\blockhdr{\emph{Active interaction}}} \\
\midrule
Qwen2.5-VL-72B & 16.8 & 26.8 & 20.8 & 19.1 & 26.7 & 46.8 & \textit{26.2} \\
Qwen3.5-9B & 36.7 & 33.5 & 21.3 & 24.7 & 30.9 & 40.1 & \textit{31.2} \\
Qwen3.5-27B & 27.5 & 40.2 & 33.2 & 31.4 & 33.8 & 33.8 & \textit{33.3} \\
InternVL3.5-38B & 30.9 & 41.5 & 43.2 & 41.6 & 50.1 & 55.8 & \textit{43.9} \\
Gemini 3.6 Flash & 43.3 & \underline{53.8} & 45.5 & \underline{44.4} & 53.4 & 73.0 & \textit{\underline{52.2}} \\
\multicolumn{8}{l}{\blockhdr{\emph{ScanQA only}}} \\
\rowcolor{oursrow} SFT init & \indom{39.4} & 46.0 & 47.3 & 41.2 & 44.7 & 68.3 & \textit{47.8} \\
\rowcolor{oursrow} Act3D (ours) & \indom{\underline{45.5}} & 49.8 & 47.6 & 42.3 & 50.7 & 67.6 & \textit{50.6} \\
\multicolumn{8}{l}{\blockhdr{\emph{Mixed data}}} \\
\rowcolor{oursrow} SFT init & \indom{39.3} & 47.4 & 39.2 & 37.7 & \indom{\underline{61.8}} & 65.9 & \textit{48.6} \\
\rowcolor{oursrow} Act3D (ours) & \indom{42.1} & 49.0 & 47.3 & 37.4 & \indom{\textbf{63.3}} & \textbf{80.0} & \textit{\textbf{53.2}} \\
\bottomrule
\end{tabular}

\caption{Performance comparison on six 3D reasoning benchmarks. We compare Act3D with existing active and passive methods. Green highlights indicate benchmark domains included in model training.}
\label{tab:leaderboard}
\vspace{-2mm}
\end{table}

As shown in Table~\ref{tab:leaderboard}, Act3D trained on mixed data achieves the highest average accuracy of 53.2\% across six benchmarks, surpassing the Qwen3.5-9B base model by 22.0\%. It further outperforms Gemini 3.6 Flash by 1.0\% under the same interaction protocol, as well as all evaluated open-source 3D reasoning methods, including both active interaction and passive perception approaches. Notably, Act3D achieves these results from a single initial view, autonomously acquiring additional visual evidence through interaction with the environment.

\textit{RL substantially improves over SFT initialization.}
In the mixed-data setting, RL improves average accuracy from 48.6\% to 53.2\% (+4.6\%) over the SFT init, with improvements on five of six benchmarks. The largest gains occur on ARKit (+14.1\%) and SPBench-MV (+8.1\%), highlighting the benefits of optimizing exploration and reasoning through verifiable RL beyond supervised trajectory imitation.

\begin{figure}[t]
\centering
\begin{minipage}[t]{0.32\linewidth}
\centering
\includegraphics[width=\linewidth]{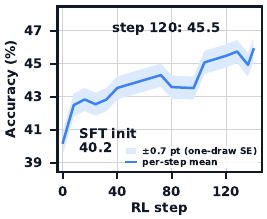}\\[-2pt]
{\small (a) ScanQA Acc. as RL training progresses (trained on ScanQA only)}
\end{minipage}\hfill
\begin{minipage}[t]{0.32\linewidth}
\centering
\includegraphics[width=\linewidth]{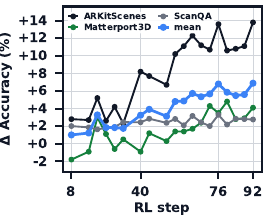}\\[-2pt]
{\small (b) $\Delta$ Acc. compared to SFT init as RL training progresses (trained on mixed data)}
\end{minipage}\hfill
\begin{minipage}[t]{0.32\linewidth}
\centering
\includegraphics[width=\linewidth]{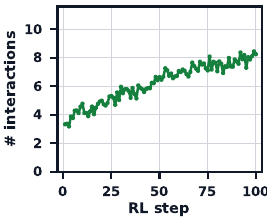}\\[-2pt]
{\small (c) Number of interactions}
\end{minipage}
\caption{\textbf{Learning dynamics of Act3D.} \emph{(a)} ScanQA-only validation.
\emph{(b)} Mixed-data validation gains relative to SFT initialization. \emph{(c)} Mean number of
interactions for the final reward. Panel (c) is in-training rollout statistics; panels (a) and (b) are validation accuracies, and no value should be read across the two.}
\label{fig:curve}
\vspace{-2mm}
\end{figure}

\textit{Improvements generalize beyond the training domain.}
To assess cross-domain generalization, we further train Act3D exclusively on ScanQA. Despite using a single training set for RL optimization, Act3D improves average accuracy from 47.8\% to 50.6\% (+2.8\%) over its SFT initialization and extends its improvement beyond ScanQA (+6.1\%), with average accuracy across the other five benchmarks increasing from 49.5\% to 51.6\% (+2.1\%). Gains are observed on four of these benchmarks, including substantial performance increases on MP3D (+6.0\%) and SQA3D (+3.8\%). These results suggest that Act3D learns transferable capabilities that generalize beyond the domain used for optimization.

\subsection{Learning Dynamics and Exploration Behavior}
\textit{RL improves reasoning performance while reshaping interaction behavior.}
Figure~\ref{fig:curve} illustrates the evolution of performance and interaction behavior during RL training. In the ScanQA-only setting, validation accuracy improves from 40.2\% to 45.5\% (Figure~\ref{fig:curve}(a)). Mixed-data training similarly yields improvements across all three validation benchmarks, with the largest gains on ARKitScenes (Figure~\ref{fig:curve}(b)). Meanwhile, the average number of interactions per episode increases from approximately 3.3 to 8.3 during mixed-data training (Figure~\ref{fig:curve}(c)), indicating that RL optimization progressively encourages more extensive interaction with the environment. These observations motivate a closer examination of how exploration contributes to reasoning performance.

\begin{figure}[h]
\centering
\includegraphics[width=0.65\linewidth]{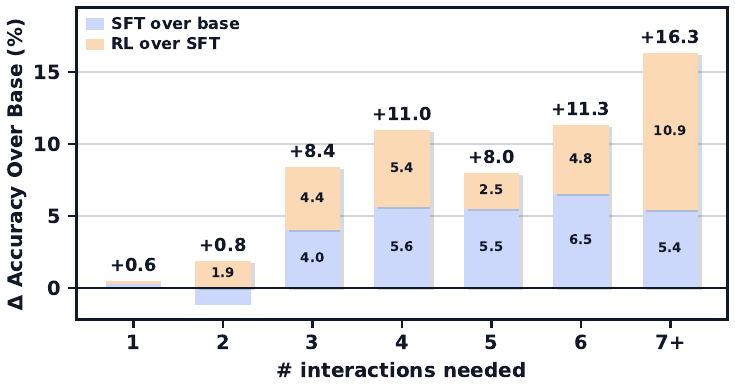}
\caption{\textbf{Performance gains across different exploration demands.} ScanQA validation questions are grouped by a geometry-based estimate of the interactions needed to reach the answer object from the initial viewpoint. Stacked bars show the accuracy improvements from SFT over the base model and subsequent RL over SFT.}
\label{fig:strata}
\vspace{-2mm}
\end{figure}

\textit{Adaptive Exploration Behavior.}
To investigate how performance gains depend on the need for exploration, we stratify ScanQA validation questions by their estimated exploration demand (Figure~\ref{fig:strata}). Specifically, we estimate the number of actions needed to approach the annotated answer object based on its bearing and distance from the initial viewpoint. This geometry-based measure is independent of the model's actual interaction behavior. We find that the combined improvements from SFT and RL are generally larger for questions requiring more exploration, increasing from 0.6\% in the lowest-demand group to 16.3\% in the highest. Notably, RL contributes an additional 10.9\% over SFT in the most demanding group. These findings suggest that the benefits of outcome-based RL are particularly pronounced when solving a question requires acquiring visual evidence beyond the initial observation.

\begin{figure}[h]
    \centering
    \includegraphics[width=1\linewidth]{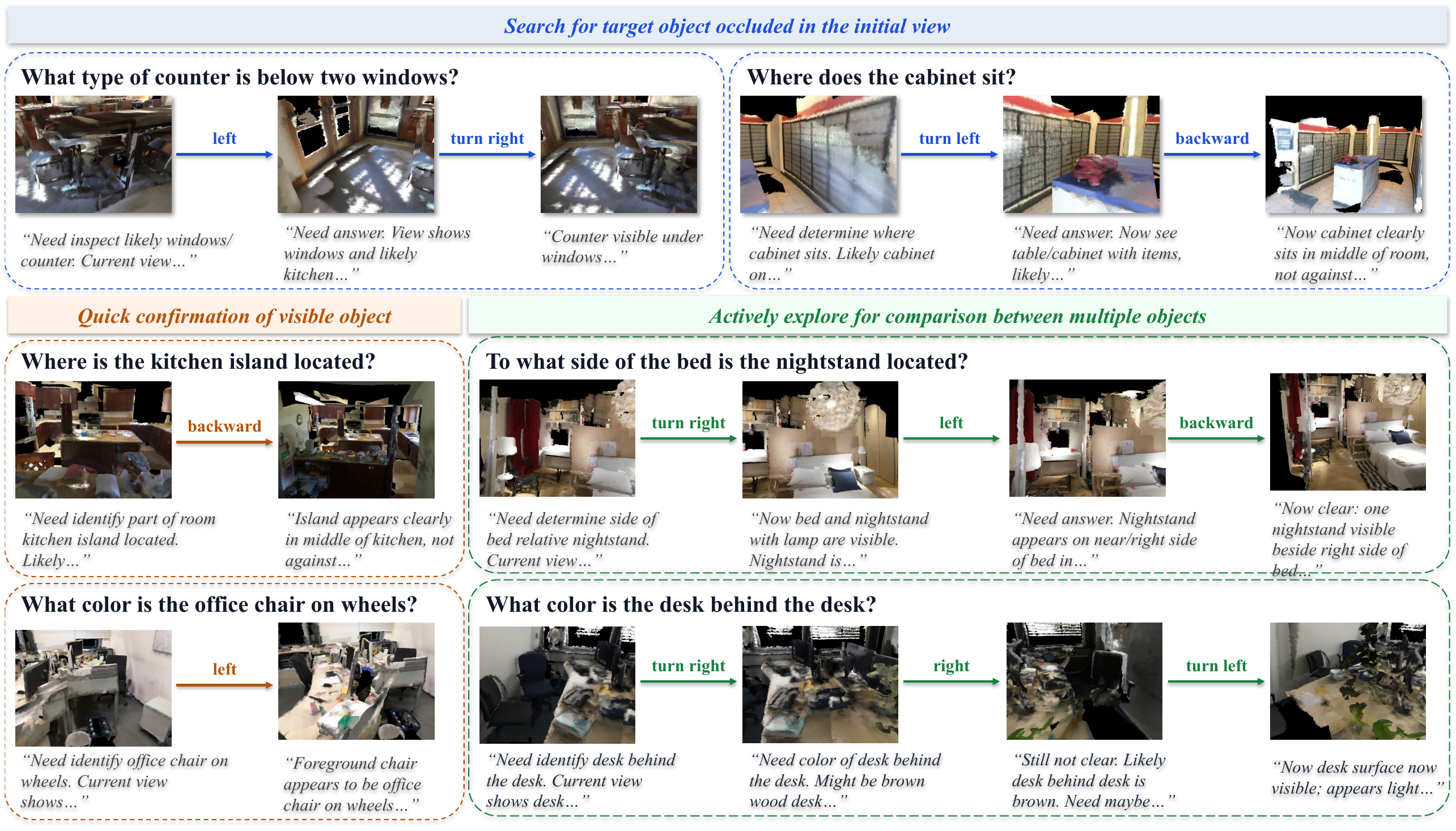}
    \caption{\textbf{Qualitative analysis.} We highlight emergent exploration behaviors. Arrows denote the executed actions, and text excerpts beneath illustrate the reasoning of the policy.}
    \label{fig:qualitative}
    \vspace{-2mm}
\end{figure}

\begin{figure}[h]
    \centering
    \includegraphics[width=1\linewidth]{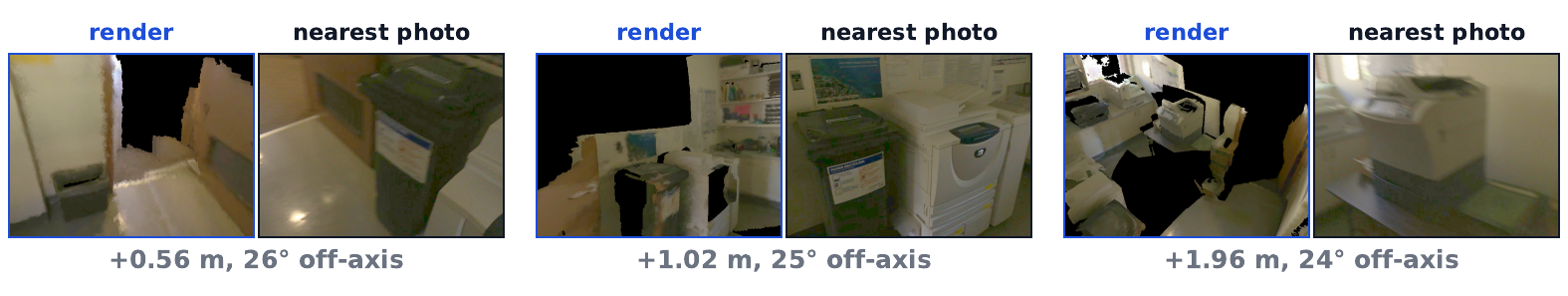}
    \caption{Rendered versus retrieved observations. Comparison of pose-aligned renders and nearest available real images from ScanNet. Distance and angular offsets indicate discrepancies between the requested and retrieved camera poses.}
    \label{fig:posematch}
\end{figure}




\subsection{Qualitative Analysis.}
Figure~\ref{fig:qualitative} illustrates several exploration strategies that emerge from Act3D's training. When relevant objects are initially occluded or outside the field of view, the policy actively changes its viewpoint to locate them before answering. When the target is already visible, it performs brief verification with minimal additional interaction. For questions requiring spatial comparisons or relationships between multiple objects, the policy explores complementary viewpoints to gather sufficient evidence. Notably, its reasoning evolves as new observations become available, allowing it to refine initial hypotheses and resolve ambiguities through successive interactions. These examples illustrate how Act3D integrates visual evidence acquisition with reasoning in a question-dependent manner.

\vspace{-1mm}
\subsection{Ablation Results}
\label{sec:ablation}
\vspace{-1mm}

\begin{wraptable}{r}{0.42\linewidth}
\vspace{-26pt}
\centering
\footnotesize
\setlength{\tabcolsep}{4pt}
\begin{tabular}{lr}
\toprule
Initial visual input & Accuracy (\%) \\
\midrule
BEV + 1 rendered frames & $36.7$ \\
BEV + 4 rendered frames & $\best{41.4}$ \\
1 rendered frame (Ours) & $\underline{41.0}$ \\
\bottomrule
\end{tabular}
\caption{ScanQA Acc. given different initial observation.}
\label{tab:bev}
\vspace{-35pt}
\end{wraptable}

\subsubsection{Initial Observation}
\label{sec:abl:bev}

We investigate how the initial visual input affects interactive reasoning performance (Table~\ref{tab:bev}). Starting from a single rendered view achieves 41.0\% accuracy on ScanQA, closely matching the 41.4\% obtained with a bird's-eye view (BEV) and four rendered views, despite using only one initial image instead of five. In contrast, combining a BEV with a single rendered view reduces accuracy to 36.7\%. These results suggest that additional initial visual context does not necessarily improve performance. We therefore adopt single-view initialization, which preserves the partial observability of the task while achieving comparable accuracy with substantially less visual input.

\subsubsection{Observation Interface: Rendering vs. Retrieval}
\label{sec:abl:input}



Act3D formulates 3D reasoning as interaction with a 3D scene representation, making the design of the visual observation interface an important consideration. Rendered observations can be generated at arbitrary camera poses but may suffer from limited visual fidelity, whereas real images preserve greater visual detail but are restricted to the viewpoints captured during data collection. We examine this trade-off by comparing pose-aligned rendering with nearest-view retrieval of real ScanNet images. As illustrated in Figure~\ref{fig:posematch}, retrieved images can deviate substantially from the requested camera poses, introducing inconsistencies between executed actions and subsequent observations. Under matched training conditions, rendering consistently outperforms real-image retrieval across all 13 evaluation checkpoints, achieving 10.6\% higher accuracy on average. These results highlight the importance of action--observation consistency in interactive 3D reasoning, even at the expense of visual fidelity.

\begin{wrapfigure}{r}{0.4\linewidth}
\centering
\includegraphics[width=\linewidth]{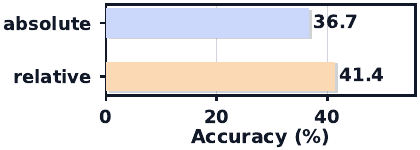}
\caption{ScanQA Acc. of different action parameterizations.}
\label{fig:actionspace}
\end{wrapfigure}

\subsection{Action Space}
\label{sec:abl:actionspace}

We investigate the effect of action parameterization by comparing relative camera movements with direct prediction of absolute camera poses. As shown in Figure~\ref{fig:actionspace}, relative actions achieve 41.4\% accuracy on ScanQA, compared with 36.7\% for absolute pose prediction. Unlike absolute coordinates, which require reasoning in a scene-dependent global reference frame, relative actions operate in the camera's local coordinate system and correspond to intuitive movements such as translation and rotation. This representation aligns more naturally with sequential, observation-conditioned exploration, motivating its adoption throughout Act3D.

\begin{figure}[t]
\centering
\includegraphics[width=\linewidth]{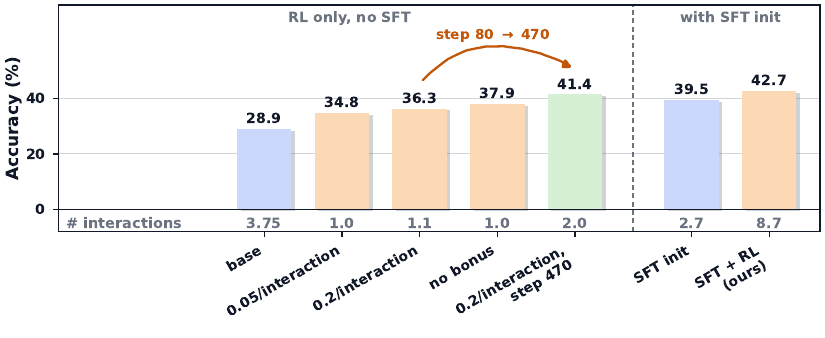}
\caption{\textbf{Effect of supervised initialization on RL.} We report ScanQA Acc. vs.\ average number of interactions per episode with and without SFT initialization. RL-only results are shown after 80 and 470 training steps.}
\label{tab:nosft}
\vspace{-4mm}
\end{figure}

\subsection{Supervised initialization}
\label{sec:abl:sft}

We investigate the role of supervised initialization in learning interactive reasoning behaviors (Figure~\ref{tab:nosft}). Without SFT, direct RL improves ScanQA accuracy from 28.9\% to 34.8--37.9\% after 80 training steps, although the policy performs only approximately one interaction per episode. Extending training to 470 steps further improves accuracy to 41.4\%, with the average number of interactions increasing to 2.0. These results suggest that RL alone can gradually improve reasoning performance while developing interaction capabilities. In contrast, SFT-initialized RL improves accuracy from 39.5\% to 42.7\%, accompanied by an increase in average number of interactions from 2.7 to 8.7 per episode. A possible explanation is that the base model is not initially adapted to the interactive environment and action protocol, whereas supervised trajectory initialization enables the model to acquire basic interaction capabilities before RL optimization. Although the settings differ in backbone and training budget, these findings suggest that SFT facilitates adaptation to the interactive setting and may accelerate subsequent RL optimization, rather than being strictly necessary for learning to interact.

\vspace{-2mm}



\subsection{Interaction Reward}
\label{sec:abl:bonus}

\begin{figure}[h]
\centering
\includegraphics[width=\linewidth]{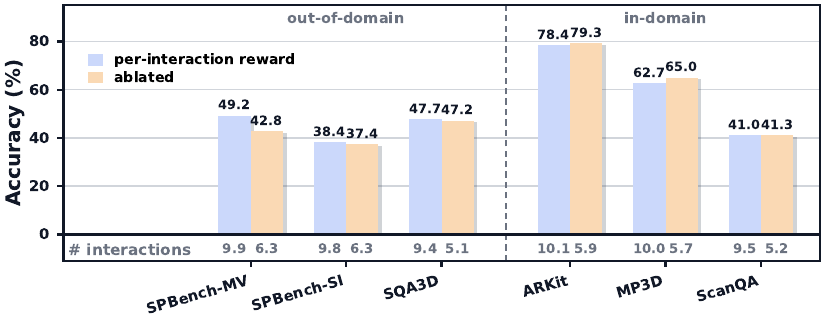}
\caption{\textbf{Effect of interaction reward.} ScanQA Acc. with and without the per-interaction reward across held-out and training benchmarks. Numbers below the bars indicate the average number of interactions per episode.}
\label{fig:bonus}
\vspace{-2mm}
\end{figure}

Explicitly rewarding interactions substantially increases exploration frequency, but does not uniformly improve reasoning accuracy (Figure~\ref{fig:bonus}). Adding a reward of 0.05 per interaction increases the average number of interactions from 5.1--6.3 to 9.4--10.1 per episode across six benchmarks. While this reward improves accuracy on all three held-out benchmarks, particularly SPBench-MV (+6.4\%), it slightly reduces performance on the training domains (0.3--2.3\%). These results suggest that encouraging additional interactions can benefit cross-domain generalization, although more extensive exploration does not necessarily yield higher accuracy. The interaction reward therefore serves primarily to encourage active evidence acquisition, while outcome-based feedback guides the policy toward effective reasoning.

\vspace{-2mm}
\section{Conclusion}

We presented Act3D, a framework for learning to think and explore in the 3D world by treating 3D representations as interactive environments rather than assumed complete inputs. Starting from a single observation, Act3D iteratively reasons, acts, and acquires visual evidence through interaction with the environment. Building on supervised trajectory initialization, verifiable RL further optimizes the policy without explicit supervision of exploration actions or viewpoints. Across six 3D benchmarks, Act3D achieves the highest average accuracy of 53.2\%, improving upon its SFT initialization by 4.6\% and outperforming existing active and passive 3D reasoning methods. Further analyses show that these gains extend beyond the training benchmark and are particularly pronounced on questions requiring more extensive exploration. Qualitative results also reveal question-conditioned exploration strategies that adapt to newly acquired observations. Together, these findings demonstrate the potential of outcome-based RL to learn not only how to reason over visual evidence, but also how to actively acquire the evidence needed for 3D reasoning.

\bibliography{iclr2027_conference}
\bibliographystyle{iclr2027_conference}

\clearpage
\appendix
\section{Why Is Interaction Necessary?}
\label{app:interaction}

We use the notation of Sec.~\ref{sec:pomdp}.
To characterize whether the observations acquired so far are sufficient to determine the answer, we define the \emph{consistency set}
\begin{equation}
\mathcal K(h_t)
=
\left\{
\mathcal M':
\operatorname{Render}(\mathcal M',c_s)=o_s,
\ \forall s\leq t
\right\},
\end{equation}
where $c_0,\ldots,c_t$ is the camera path induced by the actions in $h_t$. $\mathcal K(h_t)$ contains all scenes indistinguishable from the true scene under the interaction history $h_t$. We say that $q$ is \emph{answerable} from $h_t$ when all scenes consistent with the acquired observations induce the same answer:
\begin{equation}
g(\mathcal M',q)
=
g(\mathcal M'',q),
\qquad
\forall\,
\mathcal M',\mathcal M''
\in
\mathcal K(h_t).
\end{equation}
Because the environment is deterministic, each additional observation can only further constrain the set of compatible scenes,
\begin{equation}
\mathcal K(h_{t+1})
\subseteq
\mathcal K(h_t),
\end{equation}
and can therefore reduce uncertainty about the answer. Correspondingly, treating the interaction history as a random variable, conditioning on additional observations cannot increase the expected uncertainty about the answer:
\begin{equation}
\mathbb H(y^\star\mid h_{t+1})
\leq
\mathbb H(y^\star\mid h_t).
\end{equation}
Unlike the set inclusion, which holds along every trajectory, this inequality holds only in expectation over histories: a particular observation may leave the uncertainty unchanged or even increase it. Successful interaction therefore requires selecting viewpoints that are informative for the current question.

This formulation exposes a fundamental limitation of passive perception. A model restricted to the initial observation $o_0$ cannot recover task-relevant information absent from that view. Its Bayes-optimal passive accuracy is
\begin{equation}
\operatorname{Acc}^{\mathrm{pass}}
=
\mathbb E_{q,o_0}
\left[
\max_{y\in\mathcal Y}
\Pr(y^\star=y\mid q,o_0)
\right].
\end{equation}
Whenever $(q,o_0)$ is consistent with multiple scenes that induce different answers, this ambiguity cannot be resolved from the initial observation alone, regardless of model capacity. By contrast, interactive policies can acquire additional observations that progressively refine the set of compatible scenes. Let $\Pi_{\mathrm{int}}$ denote the class of history-dependent interactive policies over the augmented action space. Since passive prediction is recovered as the special case of terminating immediately at $t=0$,
\begin{equation}
\sup_{\pi\in\Pi_{\mathrm{int}}}
\operatorname{Acc}(\pi)
\geq
\operatorname{Acc}^{\mathrm{pass}}.
\end{equation}
Interaction can therefore overcome the passive observation ceiling by acquiring task-dependent evidence unavailable in the initial view. The resulting learning problem is not merely to reason over observed evidence, but to determine which observations should be acquired to make the question answerable.

\section{Reproducing Ross3D}
\label{app:ross3d}

\begin{table}[h]
\centering
\small
\begin{tabular}{lrr}
\toprule
metric & published & reproduced \\
\midrule
EM & 30.8 & 30.7 \\
CIDEr & 107.0 & 107.1 \\
BLEU-4 & 17.9 & 18.2 \\
METEOR & 20.9 & 20.8 \\
ROUGE-L & 50.7 & 50.7 \\
\bottomrule
\end{tabular}
\caption{Agreement with the published Ross3D results on $4{,}675$ ScanQA validation
questions.}
\label{tab:ross3dagreement}
\end{table}

We evaluate the released Ross3D checkpoint (LLaVA-Video-Qwen2-7B) with its official dataloader,
frame sampler, scorer, and bird's-eye-view input. The only implementation change is calling the
evaluation function directly because the original eight-GPU Ray launcher is incompatible with our
scheduler.

The reproduced scores closely match the published results across every metric, validating our
Ross3D evaluation pipeline and confirming that its scores are directly comparable with the other
ScanQA results reported in this paper.


\section{Prompt Templates}
\label{app:prompts}

This appendix records the prompts used by the reported policy, the supervised-data teacher,
the accuracy judge, and the proprietary baselines. Bracketed image labels denote multimodal
message parts and are not literal text sent to the model. For print consistency, em dashes are
rendered as double hyphens; all wording, placeholders, tags, and JSON structure are otherwise
preserved.

\subsection{Interactive policy and evaluation}
\label{app:policy-prompt}

The policy receives the system message and first user turn below, unchanged between
reinforcement learning and evaluation. The wording is verbatim; the sentence giving the room
dimensions is omitted when scene bounds are unavailable.

\begin{shaded}
\textbf{System.}\enspace
You are a 3D spatial reasoning assistant exploring an indoor scene. You are given ONE image:
your current camera view --- this is your starting position. To see more of the scene, call
\texttt{render\_view\_tool} with a single movement action, and optionally a distance. Actions:
\texttt{forward}, \texttt{backward}, \texttt{left}, \texttt{right}, \texttt{up}, \texttt{down},
\texttt{turn\_left}, \texttt{turn\_right}, \texttt{turn\_up}, \texttt{turn\_down}. If you do not
give a distance, the camera takes its default step --- about 1 metre forward or backward, about
0.6 metres for the other moves, and about 30 degrees for turns. Give a distance in metres (or
degrees for turns) when a larger or smaller step is more efficient --- one 4\,m move is better
than four small ones. You will see the new view after each action and can then decide to move
again or to answer. Only answer when confident; if a view is empty or low-coverage, move to a
better viewpoint first. Respond in exactly one of these two formats:
\begin{lstlisting}[style=promptbox]
<think>your reasoning here</think>
<tool_call>
{"name": "render_view_tool",
 "arguments": {"action": "<action>", "distance": <optional number>}}
</tool_call>
\end{lstlisting}
\vspace{-1pt}
\hspace*{14pt}or
\begin{lstlisting}[style=promptbox]
<think>your reasoning here</think>
<answer>your final answer</answer>
\end{lstlisting}

\vspace{6pt}
\textbf{User, turn 0.}\enspace
\tg{image}\enspace This image is your current camera view --- your starting position. The room
measures roughly \texttt{\{dx\}} m x \texttt{\{dy\}} m, with about \texttt{\{dz\}} m of height.
Move through it with \texttt{render\_view\_tool} to gather evidence, then answer.

\vspace{3pt}
\hspace*{14pt}\textbf{Question:} \texttt{\{question\}}

\vspace{3pt}
Keep the text inside the \tg{answer} tag to a SHORT PHRASE of 1--5 words --- no full sentence,
no explanation. Still reason inside \tg{think} first.
\end{shaded}

\paragraph{Tool observation.}
Every valid action returns a rendered image together with the compact observation below;
\texttt{\{ep\}} and \texttt{\{ap\}} are the camera and look-at 3-vectors rounded to 0.01 m.

\begin{center}
\includegraphics[width=\linewidth]{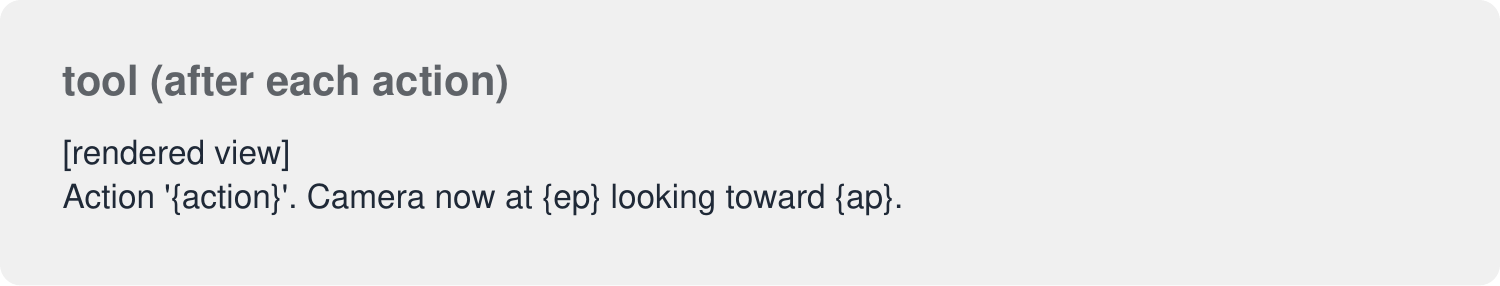}
\end{center}

\subsection{Trajectory-supervised teacher}
\label{app:sft-teacher}

The GPT-5.4 trajectory teacher reuses the policy system prompt and initial user template above.
Its per-action observation is more verbose so that the teacher can audit coverage and scene
bounds while producing the curated trajectory.

\begin{center}
\includegraphics[width=\linewidth]{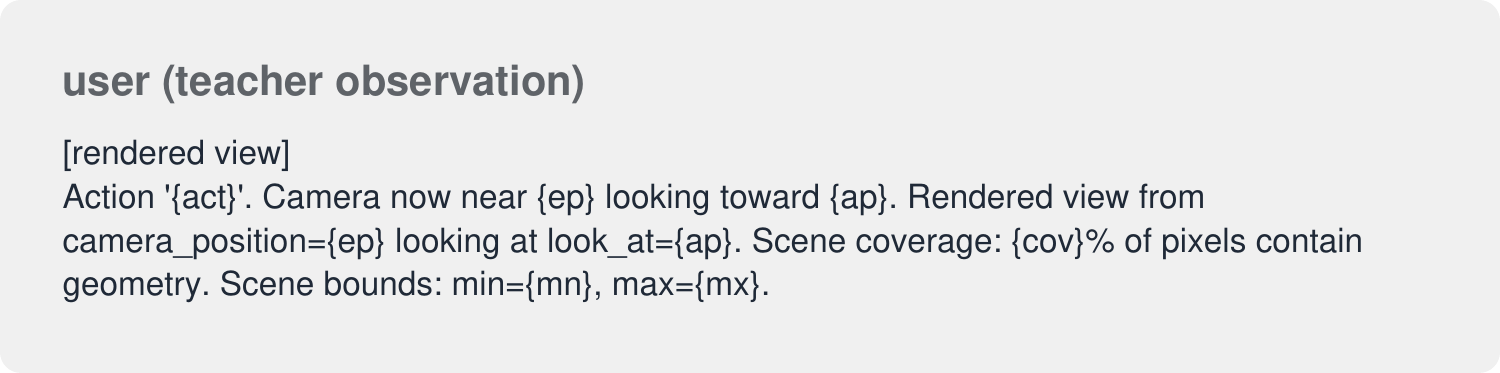}
\end{center}

The exact-match training variant uses no additional LLM prompt. It deterministically replaces
the final teacher answer with the closest ScanQA ground-truth string after normalization; exact
normalized matches take precedence over token-F1, with first-listed order breaking ties.

\subsection{Accuracy-reward judge}
\label{app:judge-prompt}

The training reward sends the following system and user messages to the local judge with
temperature 0.1 and a 64-token output limit. The binary reward is one only when the response
contains \texttt{CORRECT} and does not contain \texttt{INCORRECT}.

\begin{center}
\includegraphics[width=\linewidth]{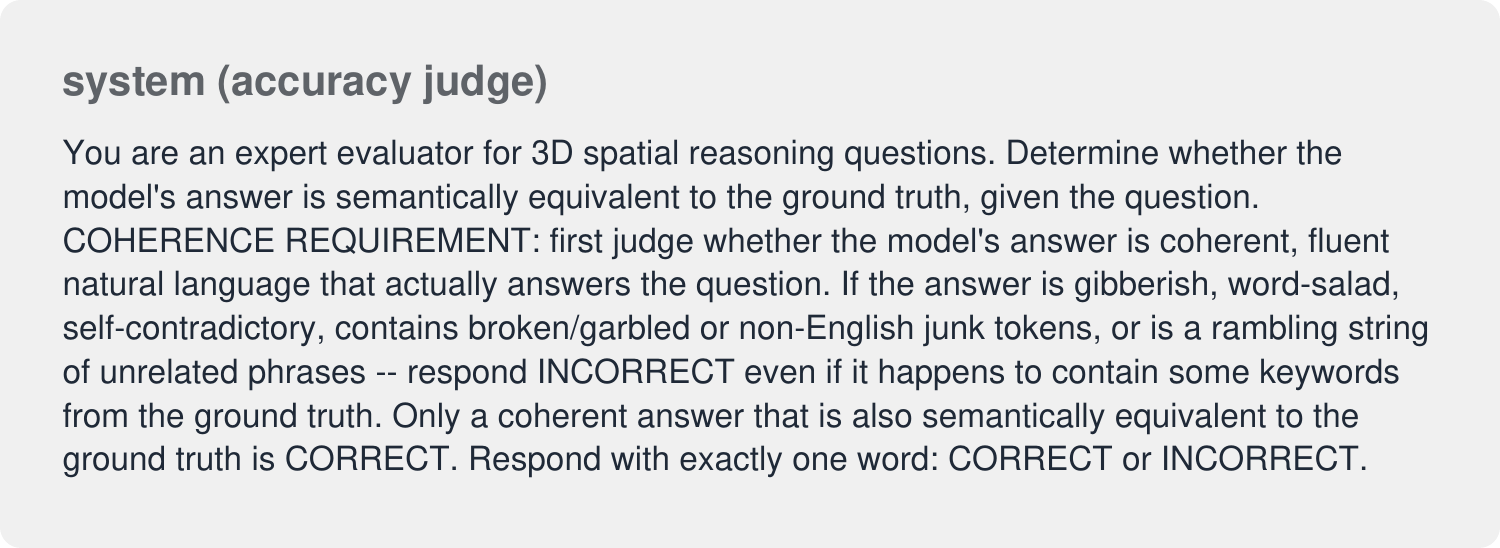}

\vspace{0.5em}
\includegraphics[width=\linewidth]{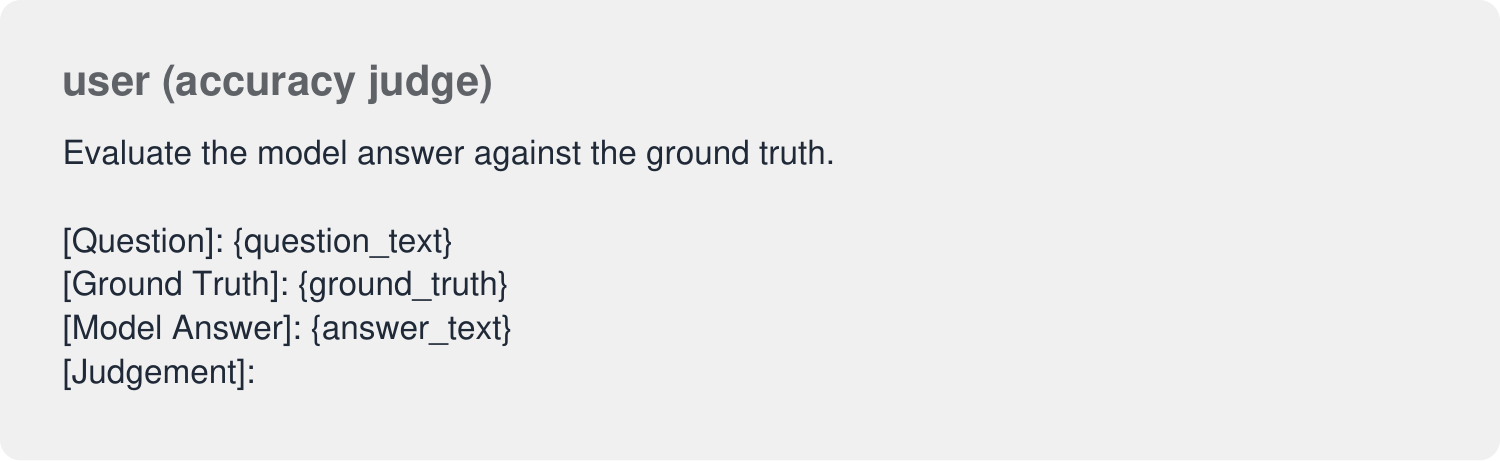}
\end{center}

\par\medskip\noindent
\begin{minipage}{\linewidth}
\paragraph{Evaluation-harness judge.}
Offline scoring of stored model predictions, including the proprietary baselines, uses a shorter
judge prompt with temperature zero and a four-token output limit. It explicitly permits terse answer fragments.

\begin{center}
\includegraphics[width=\linewidth]{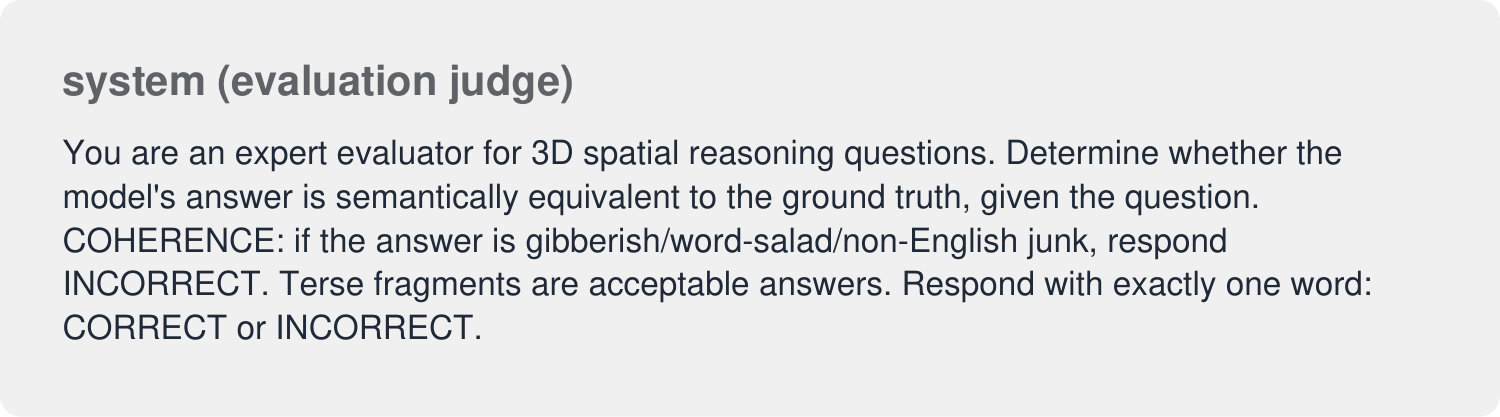}

\vspace{0.5em}
\includegraphics[width=\linewidth]{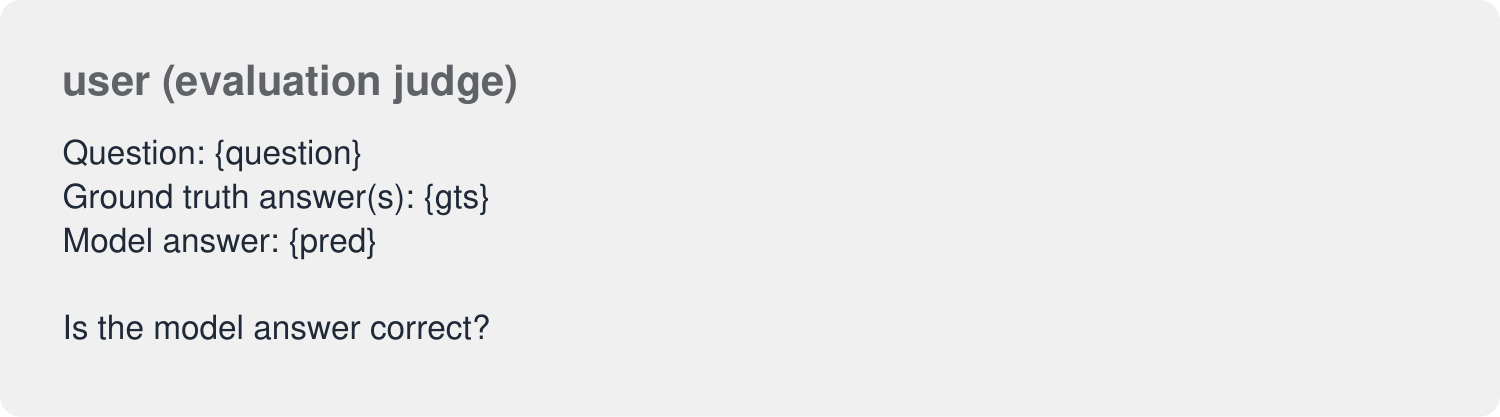}
\end{center}
\end{minipage}






\end{document}